\documentclass[10pt,twocolumn,letterpaper]{article}

\usepackage[pagenumbers]{cvpr} 

\usepackage{graphicx}
\usepackage{amsmath}
\usepackage{amssymb}
\usepackage{booktabs}
\usepackage[T1]{fontenc}

\usepackage[pagebackref,breaklinks,colorlinks]{hyperref}

\usepackage[capitalize]{cleveref}
\crefname{section}{Sec.}{Secs.}
\Crefname{section}{Section}{Sections}
\Crefname{table}{Table}{Tables}
\crefname{table}{Tab.}{Tabs.}

\begin{document}

\title{Consistent Depth Prediction under Various Illuminations using Dilated Cross Attention}

\author{Zitian Zhang\\
South China University of Technology\\
{\tt\small csjustinzhang@mail.scut.edu.cn}
\and
Chuhua Xian\\
South China University of Technology\\
{\tt\small chhxian@scut.edu.cn}
}
\maketitle

\begin{abstract}

In this paper, we aim to solve the problem of consistent depth prediction in complex scenes under various illumination conditions. The existing indoor datasets based on RGB-D sensors or virtual rendering have two critical limitations - sparse depth maps (NYU Depth V2) and non-realistic illumination (SUN CG, SceneNet RGB-D). We propose to use internet 3D indoor scenes and manually tune their illuminations to render photo-realistic RGB photos and their corresponding depth and BRDF maps, obtaining a new indoor depth dataset called Vari dataset. We propose a simple convolutional block named DCA by applying depthwise separable dilated convolution on encoded features to process global information and reduce parameters. We perform cross attention on these dilated features to retain the consistency of depth prediction under different illuminations. Our method is evaluated by comparing it with current state-of-the-art methods on Vari dataset and a significant improvement is observed in our experiments. We also conduct the ablation study, finetune our model on NYU Depth V2 and also evaluate on real-world data to further validate the effectiveness of our DCA block. The code, pre-trained weights and Vari dataset are open-sourced \footnote{\url{https://github.com/zzt76/dca}}.
 
\end{abstract}


\section{Introduction}
\label{sec:intro}

\begin{figure}[!ht]
    \centering
    \includegraphics[width=1.0\columnwidth]{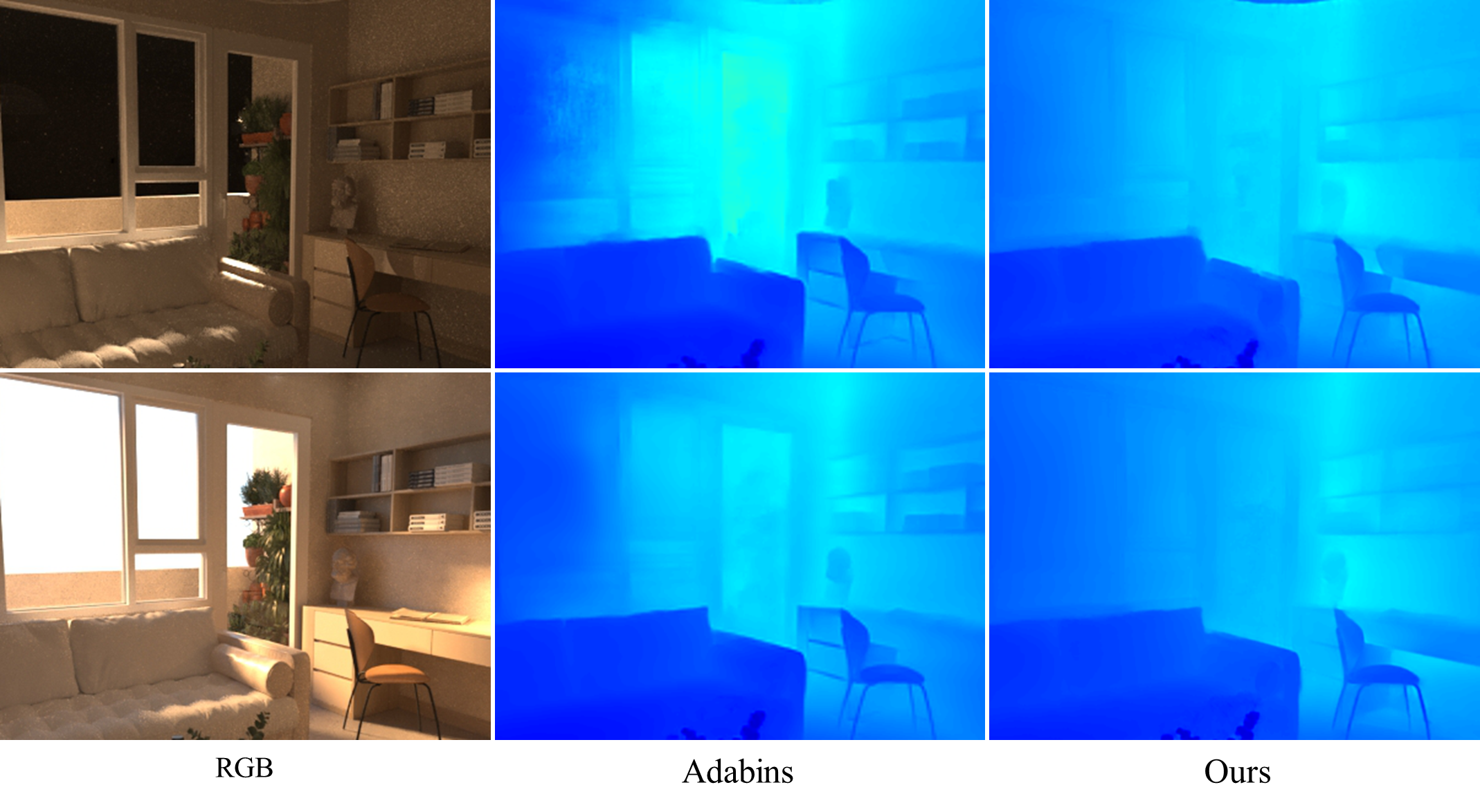}
    \caption{Comparison on the depth prediction of a scene with glass under different illuminations.}
    \label{fig:simplecomparison}
\end{figure}

Depth prediction is a classical problem in computer vision and extensively applied in robotics, auto-driving, and 3D reconstruction. People often use image stereos or videos to predict depth information and have made huge progress \cite{bregler2000recovering,faugeras2001geometry,ummenhofer2017demon,wang2019anytime}, while monocular depth prediction is still an ill-posed problem, because we can obtain the same 2D information from infinite different 3D scenes. Significant improvement have been made in this area with the development of deep learning techniques \cite{Eigen2014,laina2016deeper,qi2018geonet,bhat2021adabins}, but those methods are still sensitive to illumination changes and fail to predict the accurate depth, as \cref{fig:simplecomparison} shows.

\begin{table*}
  \centering
  \resizebox{\linewidth}{!}{
  \begin{tabular}{lcccccc}
    \toprule
     & NYUv2\cite{Silberman:ECCV12} & SUN RGB-D\cite{song2015sun} & SUN CG\cite{song2017semantic} & SceneNet RGB-D\cite{mccormac2017scenenet} & IRS\cite{wang2019irs} & Vari \\
    \midrule
    Synthetic/Natural & Natural & Natural & Synthetic & Synthetic & Synthetic & Synthetic \\
    Illumination & Real & Real & Non-realistic & Non-realistic & Non-realistic & realistic \\
    T \& R Textures & Real & Real & T & R & T \& R & T \& R \\
    Depth Map & Sparse & Sparse & Dense & Dense & Dense & Dense \\
    BRDF Maps & N & $\times$ & $\times$ & $\times$ & $\times$ & N, D, S, A \\
    \bottomrule
  \end{tabular}
  }
  \caption{The comparison of recent depth datasets and our Vari dataset. T and R are short for translucent and reflective, respectively. N, D, S, A represent surface normal, diffuse map, specular map, and albedo map, respectively.}
  \label{tab:Dataset Comparision}
\end{table*}

Different from outdoor scenes, indoor scenes have more complicated and versatile illumination conditions, spatial structures and more diverse object categories. Existing architectures often find it hard to deal with scenes they have never seen in the training set because they are not able to extract depth-relevant features and exclude irrelevant features like illumination or style from the input image. As shown in \cref{fig:simplecomparison}, it is still challenging for the state-of-the-art methods to keep consistent depth prediction while handling the same frame under different illuminations. Data augmentation does solve this problem to a certain extent by changing the brightness, contrast, gamma and rotating, cropping, flipping the picture. However, this is just a way of incrementing dataset and not changing the architecture to solve this problem. To retrieve better consistency and generalization of indoor depth prediction, we design an attention module processing encoded features to aggregate depth-relevant information.

We notice that convolutional layers only extracted global information in deeper layers while local information is discarded. Our general idea is to apply dilated convolution in each layer of the decoder meanwhile performing cross attention in these dilated features by our Dilated Cross Attention (DCA) module. It operates non-locally and produces a weighted multiplication across the features paying attention to edge, structure, and depth related information. By using DCA, we are also able to preserve feature granularity highly relevant to dense prediction by processing image at higher resolutions. Moreover, inspired by Xception~\cite{Chollet_2017_CVPR} and Yu \cite{yu2015multi}, we propose to apply depthwise separable dilated convolution (DSDC) to capture global information on higher and wider layers while the computational overhead and memory consumption remain acceptable.

We also observe current real and synthetic depth datasets have key limitations. We propose Vari Dataset, which produces photo-realistic image pairs which contains translucent and reflective surfaces and their corresponding dense depth and BRDF maps. We use Vari dataset to train and evaluate our method and compare with other state-of-the-art methods. We additionally evaluate on NYU Depth V2 and real-world data.

Our main contributions are as follows:
\begin{itemize}
    \item We propose a novel dilated cross attention (DCA) mechanism that captures global depth dependencies of scenes and aggregates depth-relevant information. We apply depthwise separable convolution during dilated operation to reduce computational consumption at high resolution.
    \item We create an indoor dataset called Vari which contains photo-realistic rendered images and corresponding dense depth map and BRDF maps, with $11$ different illuminations per frame. As far as we know, there is no such a public dataset which contains images of the same scene with different illuminations. We will release our dataset publicly.
    \item We propose a DCA-based network to predict consistent depth from a single image of the same scene under different illumination. Our method shows an impressive improvement on Vari dataset comparing to the state-of-the-art methods. Our DCA module is able to effectively predict depth from the images of complex scenes and remains consistent prediction under different illuminations. We also do an ablation study and conduct additional experiments on popular public dataset NYU Depth V2 \cite{Silberman:ECCV12} and real-world data.
\end{itemize}


\begin{figure*}[!ht]
    \centering
    \includegraphics[width=1\linewidth]{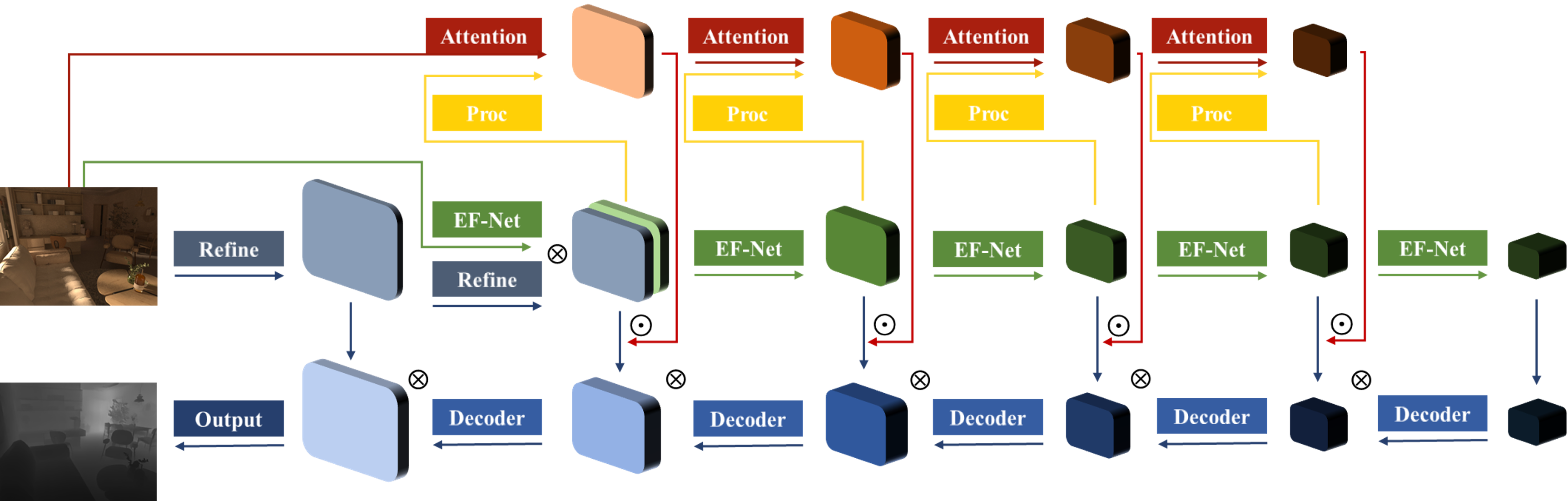}
    \caption{The framework of our proposed method. Our network consists of three major components: the EfficientNet pre-trained encoder, the feature upsampling and processing decoder, and our proposed attention branch and dilated cross attention module.}
    \label{fig:architecture}
\end{figure*}

\section{Related Work}
\label{sec:formatting}

\noindent\textbf{Monocular depth prediction.} Saxena \etal~\cite{saxena2005learning} firstly introduced a learning-based method to find a mapping from color to depth using Markov random field. Eigen \etal~\cite{Eigen2014}  proposed a coarse to fine two stage convolutional neural network to estimate depth. Since then, CNN has been wildly used for monocular depth prediction~\cite{roy2016monocular,ranftl2016dense}. Later, Laina \etal~\cite{laina2016deeper} introduced U-net \cite{ronneberger2015u} into monocular depth prediction, which was originally used for image segmentation. This encoder-decoder structure expands receptive field without significantly increasing the computational cost, which has shown great improvement handling monocular depth prediction problem~\cite{hao2018detail,fu2018deep,ramamonjisoa2019sharpnet,lee2019big,song2021monocular}. Recently, self-attention models~\cite{vaswani2017attention} dominated NLP tasks, which explicitly increase receptive field without reducing resolution. Vision transformer~\cite{dosovitskiy2020image} introduced self-attention architecture to vision tasks including depth prediction~\cite{bhat2021adabins}. Self-attention methods did achieve great success in monocular depth prediction. However, these kind of methods requires a large amount of training data to converge. Besides, self-attention aims to perform global processing of image patches, while it still lacks the ability to weight features pixel-wide. Our proposed dilated cross attention process global information and obtain depth-relevant features at the same time.

To validate our method, we consider three recent state-of-the-art methods. 
BTS~\cite{lee2019big} utilized a local planer guidance layer at each decoding stage and combine the outputs to final depth map. LapDepth~\cite{song2021monocular} merged the Laplacian pyramid into the decoder and combined the outputs of this pyramid for dense depth prediction from coarse to fine. Adabins~\cite{bhat2021adabins} performed global processing on the output of a U-Net using Vision Transformer, divided the depth range into bins and finally combined the bin center values linearly to output the depth map. 

\noindent\textbf{Indoor stereo dataset.}
Commodity RGB-D sensors such as Microsoft Kinect have been wildly used for indoor scene collection. The most popular indoor depth prediction datasets NYU Depth V2 \cite{Silberman:ECCV12}, Sun RGBD \cite{song2015sun}, Sun RGB-D are all captured by RGB-D sensors, but they still have their key limitations. These sensors failed to capture the accurate depth value of surfaces with reflective, transparent, and translucent materials such as mirrors, metals, and glasses, which are common in indoor scenes. Also, due to disparity between the infrared emitter and camera, these collected depth maps would inevitably have missing or spurious values which is fatal for dense prediction tasks, which means they may need extra manual annotation for several tasks.

Several researchers switch to train their model on synthetic images and evaluate it to real dataset directly \cite{chen2021s2r} or after fine-tuning or other training techniques \cite{ramamonjisoa2019sharpnet,Ranftl2020}, either conduct a translation or mapping from synthetic domain to real domain \cite{zheng2018t2net}. Currently, synthetic datasets such as Scene Flow \cite{MIFDB16}, IRS \cite{wang2019irs}, Scene Net RGB-D \cite{mccormac2017scenenet}, Sun CG \cite{song2017semantic} have been widely used for this purpose. \cref{tab:Dataset Comparision} compares our Vari dataset and those existing synthetic and natural datasets where we can see the limitations of previous datasets. Our Vari dataset consists of translucent and reflective objects and their precise dense depth maps, which is almost impossible for 3D sensors to capture. 

\begin{figure*}[!ht]
    \centering
    \includegraphics[width=0.95\linewidth]{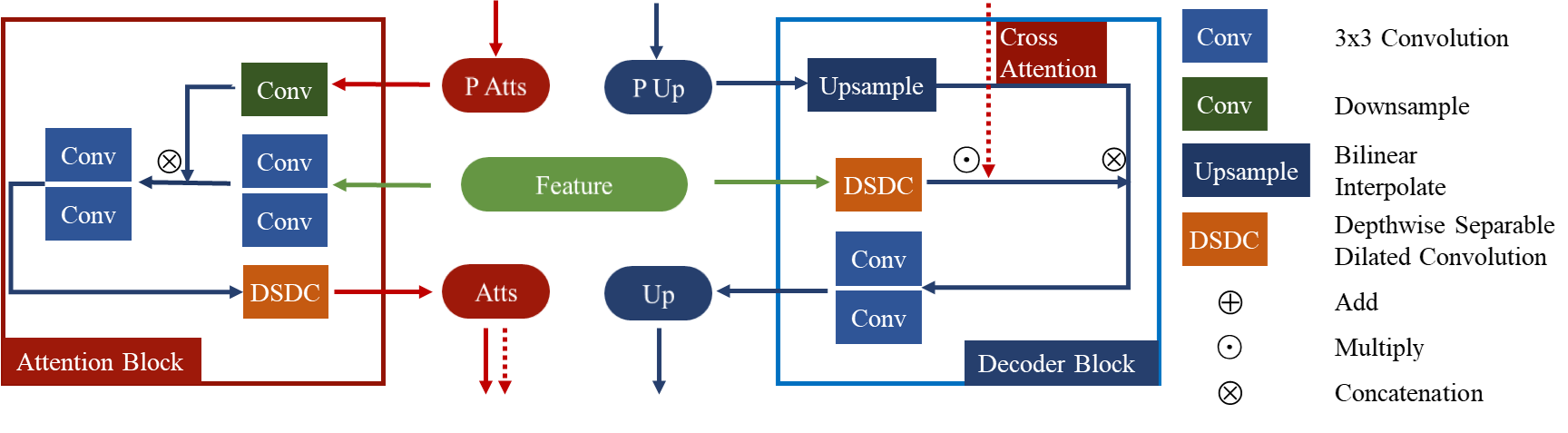}
    \caption{The illustration of dilated cross attention. The attention block takes the previous feature map and an encoded feature map as input. Then the DSDC is utilized to obtain the dilated attention map after multiple processing . In the decoder block, the DSDC is applied on the encoded feature and then multiply it with the dilated attention map pixel-wide, which called dilated cross attention.}
    \label{fig:dca}
\end{figure*}

\begin{figure}
    \centering
    \includegraphics[width=0.95\columnwidth]{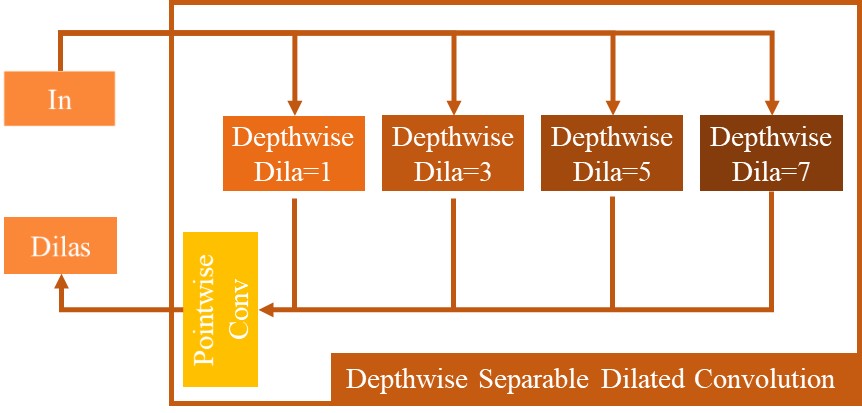}
    \caption{The depthwise separable dilated convolution.}
    \label{fig:DSDC}
\end{figure}

\section{Proposed Architecture}
\label{sec:arch}

\subsection{Overview}

The overview of our depth estimating architecture is shown in \cref{fig:architecture}.
We adopt classical U-Net \cite{ronneberger2015u} as our base architecture. Our network consists of three major components: 1) a pre-trained EfficientNet B5ap \cite{tan2019efficientnet} encoder with a refinement connection; 2) a feature upsampling and processing decoder; 3) our proposed attention branch and dilated cross attention module. We extract encoded features by downsampling and processing the input image via the encoder. We input the image to the refinement module to get full-resolution features for richer details. Then we input the image to the attention branch to obtain depth-relevant attention maps. And we conduct dilated cross attention for depth consistency and upsample the previous feature map in each state of the decoder and finally output the dense depth map.

\subsection{Details}

\noindent\textbf{Encoder}. 
We adopt EfficientNet B5ap model pre-trained on ImageNet as our backbone. Through this encoder, the input RGB image tensor $3 \times h \times w$ is downsampled and encoded as features of various dimension with channels $24,40,64,176,2048$ and referred them as encoded features. Each encoded feature map is of size $F_e \in \Re^{C_d \times \frac{h}{2^d} \times \frac{w}{2^d} }$, where $d$ is the state of encoder and $C_d$ is the current channel number. It contains concrete to abstract information for decoder and attention branch. This process could keep memory and computational consumption acceptable and allow the network to capture high-dimensional and global information. 

We also introduce a full-resolution refinement connection between the bottleneck of the encoder and decoder. We turn the input image to same-resolution features. And we concatenate them with the decoded features at the final stage to enhance local details. During both training and testing, our model takes full resolution $480\times640$ image pairs as input.

\noindent\textbf{Depthwise separable dilated convolution.} Naive convolution concerns more about local information and only processes global information in a very low resolution. We believe that global analysis done in each state of decoding is more powerful. Global processing help the network extract structural and connection information. Dilated convolution \cite{yu2015multi} adds spacing between kernel points, which exponentially expands receptive fields without losing resolution or coverage and also suppresses the noisy features. Depthwise separable convolution was first introduced by Chollet \cite{Chollet_2017_CVPR}. It decomposes standard convolution into depthwise convolutions and pointwise convolutions to learn less parameters to save memory and computational cost.

To address these problems, we propose depthwise separable dilated convolution (DSDC) module to conduct global processing and for later cross attention in this work. The structure of DSDC is shown in \cref{fig:DSDC}.  We first exploit a pyramid of dilated convolutions in depthwise to process the incoming feature globally. This module convolves the input features F separately and only in depthwise as follows:
\begin{equation}
  F_d = F \ast_d k_d,
  \label{eq:depthwise dilated convolution}
\end{equation}
where $d$ refers to dilation factors where we use $d=1,3,5,7$, and $k_d$ refers to a depthwise convolution kernel of kernel size $3\times3$ and dilation factor $d$.
Then we concatenate these four outputs $F_d, d=1,3,5,7$ into one and convolve it in point wise to aggregate the information of different receptive field and output a feature map of the same channel number and resolution as the input:
\begin{equation}
  F_p = Concat(F_d, d=1,3,5,7) \ast k_p,
  \label{eq:pointwise convolution}
\end{equation}
where $k_p$ refers to a pointwise convolution and $F_p$ is the final output. All convolutions are followed by a GELU non-linearity \cite{hendrycks2016gaussian} and batch normalization \cite{santurkar2018does}.

\noindent\textbf{Attention branch.} We noticed that the existing depth prediction network does not consider the consistency of depth prediction, and will be affected by illumination, material, or other depth-irrelevant factors, although in fact the depth will not change accordingly. Naive encoder-decoder networks are not able to deal with this problem since it cannot exclude the influence of these factor and only focus on depth-relevant factor such as structure, edge and layout. 

Our general idea is to perform a cross-attention on encoded feature maps, so that the network tend to extract depth-relevant factors and bring prediction consistency. This branch initially takes the RGB image as input. In each layer of this branch, we concatenate the output of the previous layer and a processed encoded feature map as input. Then we downsample this input of $0.5\times$ scale using convolution of kernel size $4\times4$ and stride $2$, followed by two standard convolution block of kernel size $3\times3$. Here, we apply DSDC module to obtain neighbor-aware representations and we call it dilated attention map $F_pa$. It's used for later element-wise dot product with the encoded features also processed by DSDC. 

\noindent\textbf{Decoder.} First, we apply DSDC module on the encoded feature map to produce dilated feature map. Here we operate element-wise dot product between this dilated feature map $F_{pf}$ and the dilated attention map $F_{pa}$ from the attention branch. This is what we called Dilated Cross Attention (DCA) as shown in \cref{fig:dca}:
\begin{equation}
  F = F_{pf} \bigodot F_{pa},
  \label{eq:cross attention}
\end{equation}

And we also upsample the lower-level feature map of $2\times$ scale to match the resolution of the encoded feature map $F_e$. Then we concatenate the product result with the upsampled map and operate two $3\times3$ convolutions to reduce the channel number to balance with increasing resolution. Noted that this process mostly reserves local details for dense prediction while considers the latent feature for sparse generation. The channel number of each decoder layer is $2048,1024,512,256,128$.

\subsection{Loss Function}

\noindent\textbf{Consistency loss.} We expect the estimated depth map matches the ground truth in pixel level. Hence, we train our network computing mean average error pixel-wide: 
\begin{equation}
  L_{l1} = \sum_i |y_i - y_i^*|,
  \label{eq:l1 loss}
\end{equation}
Where $y_i$ and $y_i^*$ are the valid pixels of the predicted depth map and ground truth respectively.

\noindent\textbf{Scale-invariant loss.} We adopt a scaled and square root version of Scale-Invariant loss introduced by Eigen \etal \cite{Eigen2014} to solve the depth scale ambiguity problem:
\begin{equation}
  L_{si} = \alpha \sqrt{\frac{1}{T} \sum_i g_i^2 - \frac{\lambda}{T^2} (\sum_i g_i)^2},
  \label{eq:si loss}
\end{equation}
Where $g_i = \log y_i - \log y_i^*$. $T$ refers to total number of these valid depth pixels. We set $\alpha=10, \lambda=0.85$.

\noindent\textbf{Gradient loss.} To enhance local details and the depth discontinuities of edges and boundaries in the predicted depth maps, we use the gradient loss proposed by Song \etal \cite{song2021monocular}:
\begin{equation}
  L_{grad} = \frac{1}{T} \sum_i |y_{i,h} - y_{i,h}^*| + |y_{i,w} - y_{i,w}^*|,
  \label{eq:grad loss}
\end{equation}
where $y_{i,h}$ and $y_{i,h}^*$ denote the gradient values of the predicted depth and ground truth in the horizontal direction respectively, $y_{i,w}$ and $y_{i,w}^*$ denote the gradient values in the vertical direction.

\noindent\textbf{Total loss.} The losses are weighted and added to the total loss $L_{total}$:
\begin{equation}
  L_{total} = \lambda_1 L_{l1} + \lambda_2 L_{si} + \lambda_3 L_{grad},
  \label{eq:total loss}
\end{equation}
where $\lambda_1, \lambda_2, \lambda_3$ are $0,1,0.1$ when training on Vari dataset and $1,0.02,0.1$ when training on NYU Depth V2 dataset respectively.


\section{Experiment}

We first introduce our dataset of complex indoor scenes with various and realistic illuminations called the Vari dataset. We trained and evaluated our network and the state-of-the-art methods on Vari dataset for comparing the performance and consistency, and then conduct an ablation study. We also evaluate our method on popular indoor dataset NYU Depth V2 \cite{Silberman:ECCV12}.

\subsection{The Vari Dataset}

\begin{figure}
    \centering
    \includegraphics[width=0.95\linewidth]{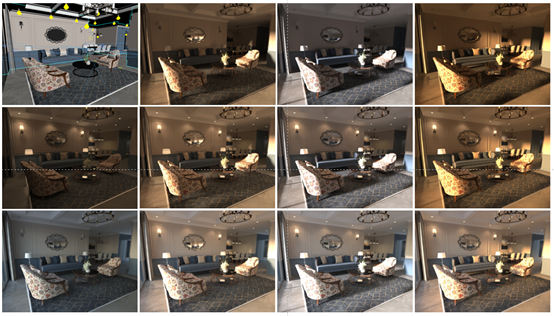}
    \caption{An example of the rendered frame of $11$ different illuminations. The left top is the original scene and the rest is ordered as listed in \cref{tab:Illumination Combinations}.}
    \label{fig:Vari example}
\end{figure}

\begin{table}
  \centering
  \resizebox{\linewidth}{!}{
  \begin{tabular}{l|ccc}
    \toprule 
     Illuminations &  SunMorning(M) & SunNoon(Nn) & SunNight(Nt) \\ 
     \midrule
     Indoor(I) & M+I & Nn+I & Nt+I \\
     Environment(E) & M+I+E & Nn+I+E & Nt+I+E \\
    \bottomrule
  \end{tabular}}
  \caption{Illumination Combinations.}
  \label{tab:Illumination Combinations}
\end{table}

\begin{figure}
  \centering
   \includegraphics[width=0.95\linewidth]{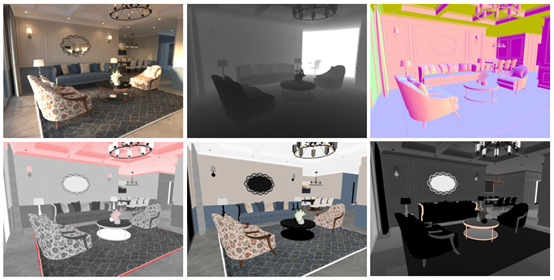}
   \caption{An example of the image pair in our Vari dataset. Sequentially, they are the RGB image, dense depth map, surface normal map, albedo map, diffuse map,  and specular map.}
   \label{fig:brdf maps}
\end{figure}

\begin{figure*}
    \centering
    \includegraphics[width=1\linewidth]{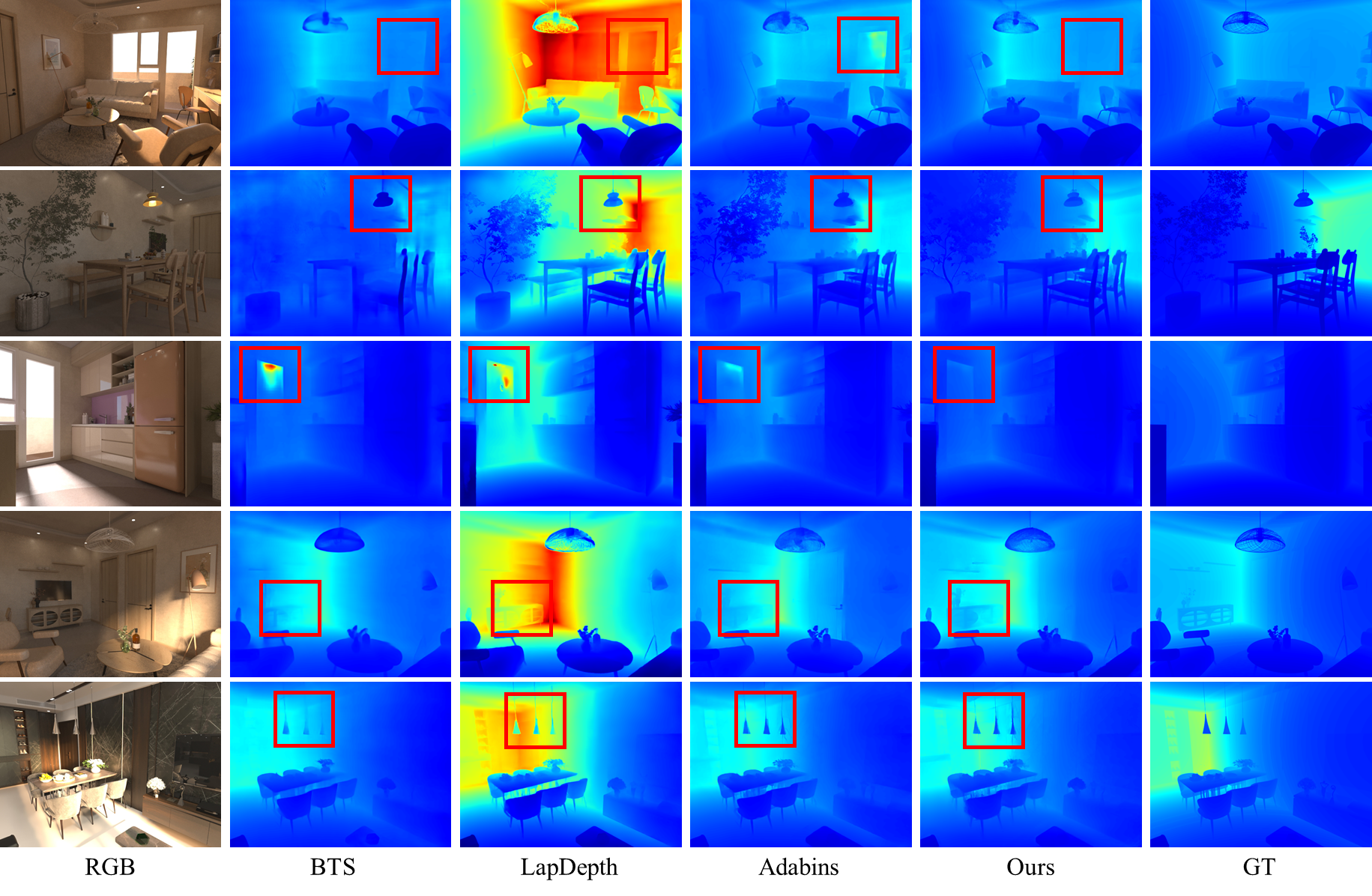}
    \caption{Qualitative comparison with the state-of-the-art methods on Vari dataset.}
    \label{fig:Results on Vari}
\end{figure*}

To evaluate the consistency of the depth prediction of our method under different illumination, we construct a new dataset named Vari with various illuminations of the indoor scenes, and it's publicly available. 
This dataset is generated by 3ds Max with Corona Renderer with various customized lighting conditions. With the path tracing techniques of this renderer, we can well simulate the real-world vision effect of indoor scenes, and furthermore produce synchronized geometry and rendering parameters of the scenes.

We first download hand-craft complex indoor 3ds Max scenes from \href{https://3dzip.org/}{3dzip.org}. In these scenes, there are large areas of high glossiness objects, such as metal, glass, mirror, and many translucent materials, e.g. glass, plastic, or jewel, which are challenging to handle for current depth prediction methods. 


Before rendering, we adjust the scene layout, surface material, and other rendering parameters to generate better visual effects. Then, we group the original indoor lights together, add ambient light and $3$ different sun lights of morning, noon, and afternoon to the the scenes. We arrange these $3$ kinds of lights in a certain way so that for each frame.~\cref{tab:Illumination Combinations} lists the illumination combinations used in rendering. There are $11$ different illuminations in the same viewpoint. ~\cref{fig:Vari example} shows an example of one rendered frame. In our depth prediction task in this paper, all these $11$ are used for training.     


When these scenes was rendering, the Corona Renderer was configured to compute the geometry parameters such as the world normal and the depth relatively to the camera. In addition, the bidirectional reflectance distribution function (BRDF) parameters (diffuse, specular, albedo, normal) were also generated and recorded. ~\cref{fig:brdf maps} shows an example of the image pair of each frame in our dataset. Thus, our dataset can be also applied to other supervised learning task, such as differential rendering, not limited in depth prediction.     

We adjust the virtual camera and shoot across the scene. Finally,  we generated more than $50$k pairs of $480 \times 640$ resolution color maps on different illuminations with $5$ geometry and BRDF maps (among $51658$ for training and $2992$ for testing) from $47$ unique indoor scenes. 

\begin{figure*}
    \centering
    \includegraphics[width=1\linewidth]{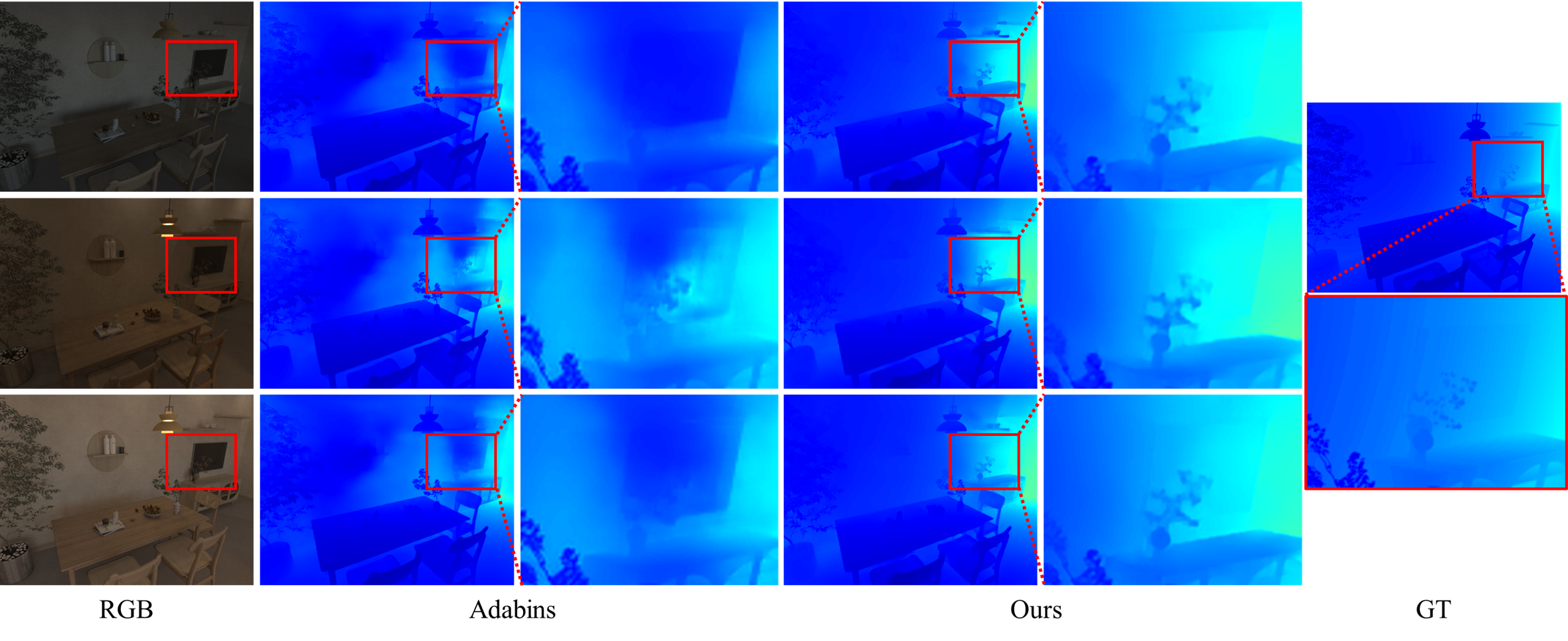}
    \caption{An example of comparison with Adabins~\cite{bhat2021adabins} under $3$ different illuminations.}
    \label{fig:Consistency Evaluation}
\end{figure*}

\subsection{Implementation Details}

We implemented our proposed method in PyTorch \cite{paszke2019pytorch}. The experiments are conducted on a PC with $3.0$GHz Intel Core i7 processor and two NVIDIA GeForce RTX 3090 $24$GB GPUs. For training, we set the total number of epochs of $25$ with batch size $6$. We use AdamW optimizer \cite{kingma2014adam,loshchilov2017decoupled} and learning rate is set to $10^{-4}$. And we adjust the learning rate after every epoch by using Exponential Learning Rate Scheduler of learning rate decay rate $0.97$. We adopt the EfficientNet B5ap \cite{tan2019efficientnet} as the encoder for feature extraction whose parameters are pre-trained on ImageNet. Our model has roughly $75$M parameters, among $28$M for the EfficientNet encoder, $45.3$M for the CNN decoder and $1.7$M for our DCA module.

During training, we perform the online data augmentation to improve robustness and generalization of our model. Specifically, we first rotate the input images by $-2.5~2.5$ degree, and randomly crop them from original resolution $480 \times 640$ to $416\times 544$. We also flip the input pairs horizontally and adjust the gamma, brightness, contrast, and sharpness of the RGB image. In the test time, we simultaneously input the RGB image and its horizontally flipped one into our model and compute the average of the original output and the flipped one as the final output.

\subsection{Evaluation Metrics}
We follow previous works and use the same six metrics introduced by Eigen \etal~\cite{Eigen2014}. These metrics are defined as: 

\noindent Threshold accuracy ($\delta$): $\%$ of $y_i$ s.t. $max(\frac{y}{y*},\frac{y*}{y})=\delta<th$ for $th=1.25,1.25^2,1.25^3$; 

\noindent Average relative error: $\frac{1}{n} \sum_{n} \frac{|y-y*|}{y} $; 

\noindent Root mean squared error (RMS): $\sqrt{\frac{1}{n}\sum_{n}(y-y*)^2}$; 

\noindent Average $log_{10}$ error: $\frac{1}{n} \sum_n |{log}_{10}(y)-{log}_{10}(y*)|$, where $y$ and $y*$ denote the pixels of predicted depth map and ground truth respectively, and $n$ is the total number of valid pixels in the ground truth.

\subsection{Evaluations}

\noindent\textbf{Comparisons on Vari dataset.} We compare our proposed method with the leading monocular depth prediction models, BTS \cite{lee2019big}, LapDepth \cite{song2021monocular} and the current State-of-the-art method Adabins \cite{bhat2021adabins} on NYU Depth V2 dataset \cite{Silberman:ECCV12}. We follow their training configurations to retrain these models on our Vari dataset using their authors code. We evaluate them using the evaluation metrics described above. 

\begin{table}
  \centering
  \resizebox{\linewidth}{!}{
  \begin{tabular}{lcccccc}
    \toprule
     & $\delta_1 \uparrow$  & $\delta_2 \uparrow$  & $\delta_3 \uparrow$  & AbsRel $\downarrow$ & RMSE $\downarrow$ & $log_{10} \downarrow$ \\
    \midrule
    BTS \cite{lee2019big} & 0.720 & 0.913 & 0.966 & 0.192 & 0.500 & 0.081 \\
    Adabins \cite{bhat2021adabins} & 0.770 & 0.952 & 0.984 & 0.176 & \underline{0.404} & 0.068 \\
    LapDepth \cite{song2021monocular} & \underline{0.788} & \underline{0.955} & \underline{0.986} & \underline{0.170} & 0.405 & \underline{0.066} \\
    \midrule
    \textbf{Ours} & \textbf{0.797} & \textbf{0.956} & \textbf{0.988} & \textbf{0.161} & \textbf{0.377} & \textbf{0.063} \\
    \bottomrule
  \end{tabular}}
  \caption{Comparison of performances on Vari dataset. The best results are in bold, the second best results are underlined.}
  \label{tab:Comparison on Vari}
\end{table}

The qualitative result is shown in \cref{fig:Results on Vari} comparing with the state-of-the-art methods, which illustrates the performance of our model and the limitations of previous methods. Specifically, though the depth maps predicted by Adabins \cite{bhat2021adabins} and our method are generally accurate, BTS \cite{lee2019big} failed to handle the place where the depth value changes smoothly. And LapDepth \cite{song2021monocular} predicts in the wrong range though the relative depth relationship is mostly right and perform well on the evaluation metrics. As for details, previous methods yield unexpected prediction on translucent areas such as glasses and suffer from granularity loss, while our method performs better apparently.

We especially compare the consistency with the state-of-the-art method Adabins. As shown in \cref{fig:Consistency Evaluation}, Adabins predicts an large area of wrong depth values on the middle of the wall, where the depth values should be increasing smoothly. For convenience, we use a red box to mark the place that we should focus and zoom in to compare the details. Adabins is disturbed by the hanging television and yields weird and unexpected results where the depth values are close the the wall. And it isn't able to separate the potted plant from background. In terms of consistency, estimated depth values by Adabins change significantly with the illuminations. Our method outperforms Adabins remarkably in both accuracy and consistency, which we believes our proposed architecture plays an important role.

\begin{figure}
  \centering
   \includegraphics[width=0.95\linewidth]{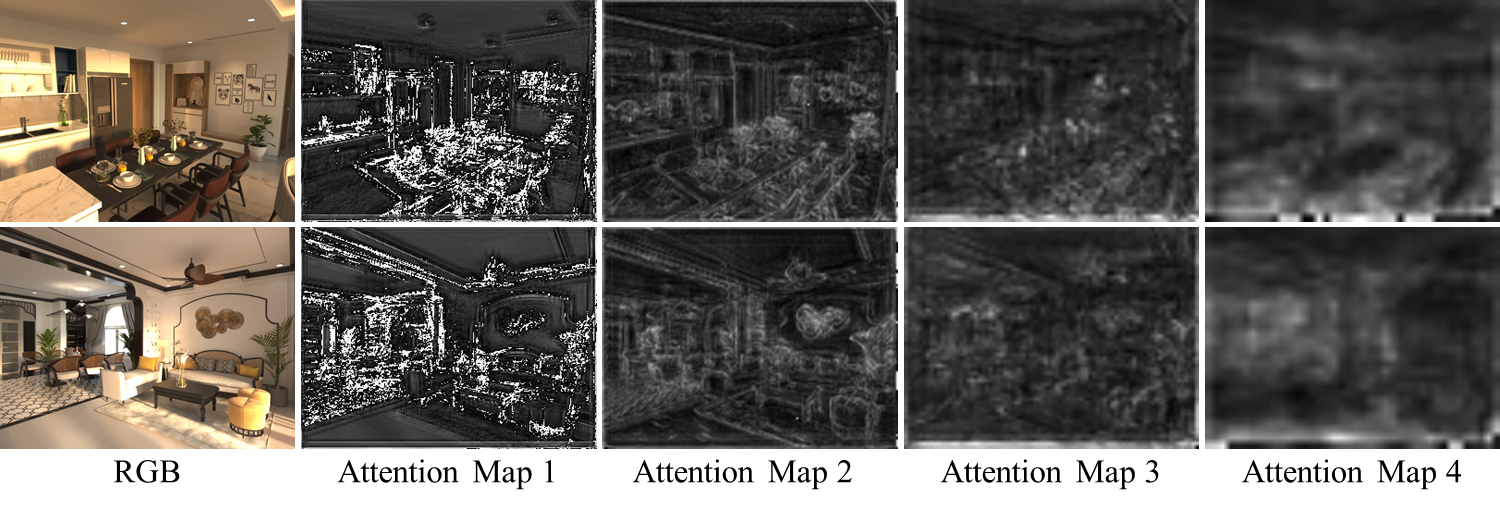}
   \caption{Visualization of the dilated attention maps.}
   \label{fig:attention maps}
\end{figure}

We also visualize the dilated attention maps in \cref{fig:attention maps} by taking the mean of them. These maps outline the depth-relevant features such as boundaries, edges, which helps to maintain the depth consistency. And it surpasses the surfaces where the depth value is uniformly distributed and focuses on where the depth value changes drastically for better and more detailed prediction.

\noindent\textbf{Ablation study.} We evaluate our DCA module by doing ablation study. We removed the attention branch, depthwise separable dialted convolution block and the cross attention operation and use the native encoder-decoder to predict the depth map. \cref{tab:Ablation study} shows the comparison, which indicates that the model with our DCA module performs better than the base model.

\begin{table}
  \centering
  \resizebox{\linewidth}{!}{
  \begin{tabular}{l c c c c c c}
    \toprule
     & $\delta_1 \uparrow$  & $\delta_2 \uparrow$  & $\delta_3 \uparrow$  & AbsRel $\downarrow$ & RMSE $\downarrow$ & $log_{10} \downarrow$ \\
    \midrule
    Base & 0.777 & 0.950 & 0.986 & 0.174 & 0.387 & 0.067 \\
    \textbf{Base + DCA} & \textbf{0.797} & \textbf{0.956} & \textbf{0.988} & \textbf{0.161} & \textbf{0.377} & \textbf{0.063} \\
    \bottomrule
  \end{tabular}}
  \caption{Numeric data of ablation study. The best results are in bold.}
  \label{tab:Ablation study}
\end{table}

\subsection{Generalization Study}

\noindent\textbf{Generalization on public datasets.} The NYU Depth V2 dataset \cite{Silberman:ECCV12} is comprised of RGB and depth video sequences of 464 indoor scenes with resolution of $640\times480$ recorded by Microsoft Kinect. The depth values range from 0 to 10 meters. We train our network on a 24K subset and evaluate on 1449 labeled pairs which are center cropped pre-defined by Eigen \cite{Eigen2014} to avoid blank edges. Then we further do Eigen crop to the image to $416\times544$, and follow the data augmentation configurations used on Vari dataset. Our network final outputs the same resolution of the input.

We finetune the model pre-trained on Vari dataset and it predicts accurate depth on NYU Depth V2 dataset. However, as the results shown in \cref{tab:Comparison on NYU Depth V2}, Our method still has several limitations. Architecture based on convolution fails to capture as much global-information as self-attention methods. And our DCA module is not efficient enough to extract depth-relevant features and not able to capture enough global information from sparse depth dataset. Because there are average $32.5\%$ missing values in the ground truth of NYU Depth V2 where the missing values often appear in edges which is important for depth prediction. And the depth values of highly reflective and translucent objects are missing or wrong because the infrared ray of the depth sensor may directly pass through the translucent or transparent object or be reflected by the specular surfaces. It may lead to false ground truth where our method pre-trained on Vari dataset could predict the accurate depth of translucent and reflective objects.

\begin{table}
  \centering
  \resizebox{\linewidth}{!}{
  \begin{tabular}{lcccccc}
    \toprule
     & $\delta_1 \uparrow$  & $\delta_2 \uparrow$  & $\delta_3 \uparrow$  & AbsRel $\downarrow$ & RMSE $\downarrow$ & $log_{10} \downarrow$ \\
    \midrule
    Eigen \etal \cite{Eigen2014} & 0.769 & 0.950 & 0.988 & 0.158 & 0.641 & 0.095 \\
    Laina \etal \cite{laina2016deeper} & 0.811 & 0.953 & 0.988 & 0.127 & 0.573 & 0.055 \\
    Geonet \cite{qi2018geonet} & 0.834 & 0.960 & 0.990 & 0.128 & 0.569 & 0.057 \\ 
    Fu \etal \cite{fu2018deep} & 0.828 & 0.965 & 0.992 & 0.115 & 0.509 & 0.051 \\
    SharpNet \cite{ramamonjisoa2019sharpnet} & 0.836 & 0.966 & 0.993 & 0.139 & 0.502 & \underline{0.047} \\
    VNL \cite{yin2019enforcing} & 0.875 & 0.976 & 0.994 & 0.111 & 0.416 & 0.048 \\
    DAV \cite{huynh2020guiding} & 0.882 & \underline{0.980} & \underline{0.996} & \underline{0.108} & 0.412 & - \\
    LapDepth \cite{song2021monocular} & \underline{0.885} & 0.979 & 0.995 & 0.110 & 0.393 & \underline{0.047} \\
    BTS \cite{lee2019big} & \underline{0.885} & 0.978 & 0.994 & 0.110 & \underline{0.392} & \underline{0.047} \\
    AdaBins \cite{bhat2021adabins} & \textbf{0.903} & \textbf{0.984} & \textbf{0.997} & \textbf{0.103} & \textbf{0.364} & \textbf{0.044} \\
    \midrule
    \textbf{Ours} & 0.865 & 0.976 & \underline{0.996} & 0.119 & 0.400 & 0.050 \\
    \bottomrule
  \end{tabular}}
  \caption{Comparison of performances on NYU Depth V2 dataset. The best results are in bold, the second best results are underlined.}
  \label{tab:Comparison on NYU Depth V2}
\end{table}

\begin{figure}
  \centering
   \includegraphics[width=1\linewidth]{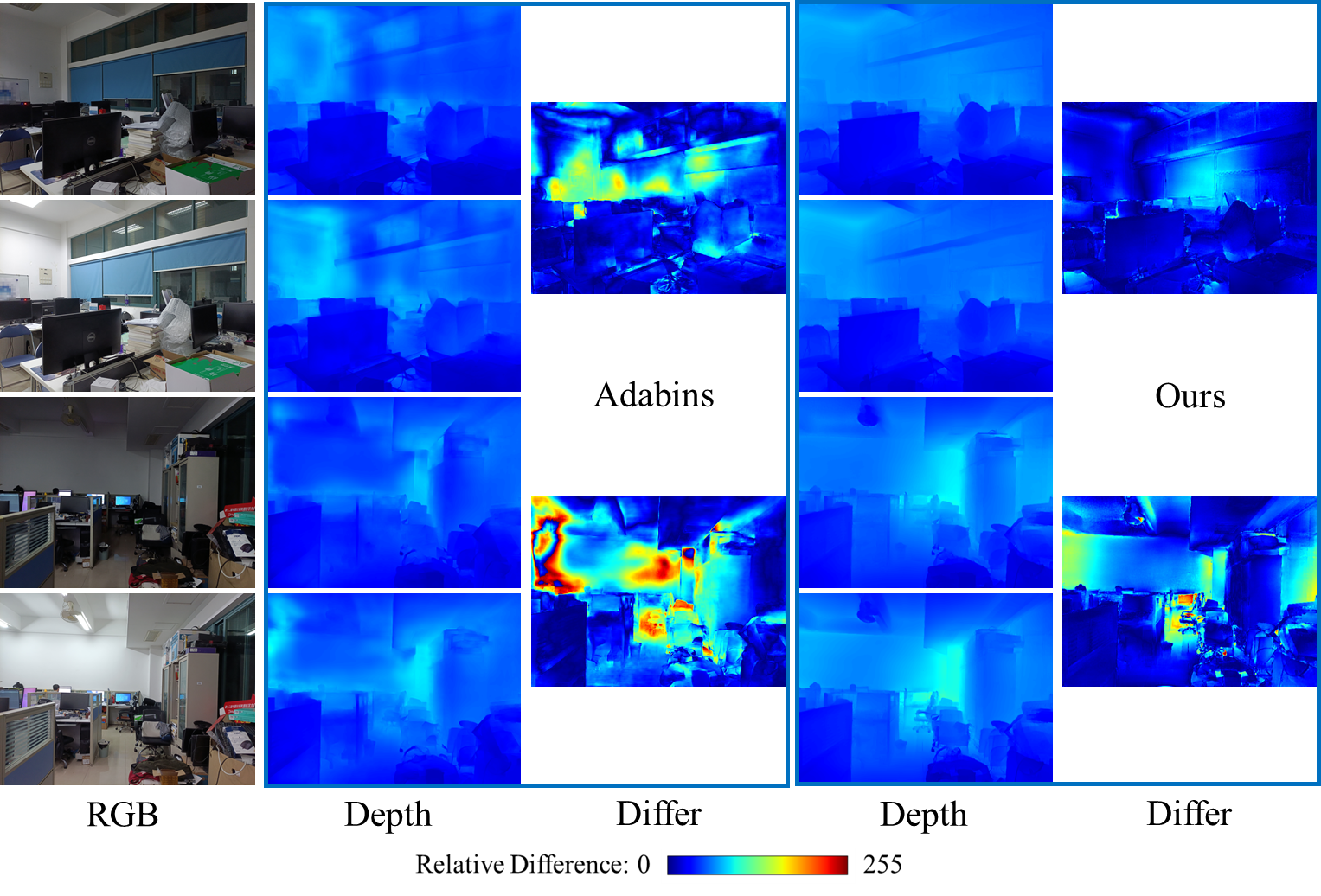}
   \caption{Comparisons on real-world data. The difference map illustrates the relative difference between depth maps under two illuminations.}
   \label{fig:real data}
\end{figure}

\noindent\textbf{Generalization in real scenes.}
To validate the generalization in real scenes, we use mobile phone to capture several photos in the same scene from the same viewpoint under different illuminations and conduct comparisons with Adabins \cite{bhat2021adabins}.
As shown in \cref{fig:real data}, our method is able to predict reliable and consistent depth on real-world data while Adabins estimates blurry and unexpected depth values. Our method outperforms Adabins in both performance and consistency, and this huge gap also emphasises that our DCA module has a huge impact on the generalization of our model.


\section{Conclusion}
\label{sec:conclusion}

In this paper, we introduce a novel building block called DCA module to perform global processing and cross attention during decoding and propose a DCA-based depth estimation model. It predicts precise and consistent depth of the same frame under different illuminations. To evaluate the effectiveness of our method, we build a novel dataset called Vari, where each frame contains a rendered realistic image, a corresponding depth map, and $4$ BRDF maps. Experimental results demonstrate that our method outperforms the state-of-the-art methods on both Vari dataset and real-world data. In the future, we plan to extend our work to inverse rendering tasks based on Vari dataset. 

{\small
\bibliographystyle{ieee_fullname}
\bibliography{dca}

\begin{thebibliography}{10}\itemsep=-1pt

\bibitem{bhat2021adabins}
Shariq~Farooq Bhat, Ibraheem Alhashim, and Peter Wonka.
\newblock Adabins: Depth estimation using adaptive bins.
\newblock In {\em Proceedings of the IEEE/CVF Conference on Computer Vision and
  Pattern Recognition}, pages 4009--4018, 2021.

\bibitem{bregler2000recovering}
Christoph Bregler, Aaron Hertzmann, and Henning Biermann.
\newblock Recovering non-rigid 3d shape from image streams.
\newblock In {\em Proceedings IEEE Conference on Computer Vision and Pattern
  Recognition. CVPR 2000 (Cat. No. PR00662)}, volume~2, pages 690--696. IEEE,
  2000.

\bibitem{chen2021s2r}
Xiaotian Chen, Yuwang Wang, Xuejin Chen, and Wenjun Zeng.
\newblock S2r-depthnet: Learning a generalizable depth-specific structural
  representation.
\newblock In {\em Proceedings of the IEEE/CVF Conference on Computer Vision and
  Pattern Recognition}, pages 3034--3043, 2021.

\bibitem{Chollet_2017_CVPR}
Francois Chollet.
\newblock Xception: Deep learning with depthwise separable convolutions.
\newblock In {\em Proceedings of the IEEE Conference on Computer Vision and
  Pattern Recognition (CVPR)}, July 2017.

\bibitem{dosovitskiy2020image}
Alexey Dosovitskiy, Lucas Beyer, Alexander Kolesnikov, Dirk Weissenborn,
  Xiaohua Zhai, Thomas Unterthiner, Mostafa Dehghani, Matthias Minderer, Georg
  Heigold, Sylvain Gelly, et~al.
\newblock An image is worth 16x16 words: Transformers for image recognition at
  scale.
\newblock {\em arXiv preprint arXiv:2010.11929}, 2020.

\bibitem{Eigen2014}
David Eigen, Christian Puhrsch, and Rob Fergus.
\newblock Depth map prediction from a single image using a multi-scale deep
  network.
\newblock {\em arXiv preprint arXiv:1406.2283}, 2014.

\bibitem{faugeras2001geometry}
Olivier Faugeras and Quang-Tuan Luong.
\newblock {\em The geometry of multiple images: the laws that govern the
  formation of multiple images of a scene and some of their applications}.
\newblock MIT press, 2001.

\bibitem{fu2018deep}
Huan Fu, Mingming Gong, Chaohui Wang, Kayhan Batmanghelich, and Dacheng Tao.
\newblock Deep ordinal regression network for monocular depth estimation.
\newblock In {\em Proceedings of the IEEE conference on computer vision and
  pattern recognition}, pages 2002--2011, 2018.

\bibitem{hao2018detail}
Zhixiang Hao, Yu Li, Shaodi You, and Feng Lu.
\newblock Detail preserving depth estimation from a single image using
  attention guided networks.
\newblock In {\em 2018 International Conference on 3D Vision (3DV)}, pages
  304--313. IEEE, 2018.

\bibitem{hendrycks2016gaussian}
Dan Hendrycks and Kevin Gimpel.
\newblock Gaussian error linear units (gelus).
\newblock {\em arXiv preprint arXiv:1606.08415}, 2016.

\bibitem{huynh2020guiding}
Lam Huynh, Phong Nguyen-Ha, Jiri Matas, Esa Rahtu, and Janne Heikkil{\"a}.
\newblock Guiding monocular depth estimation using depth-attention volume.
\newblock In {\em European Conference on Computer Vision}, pages 581--597.
  Springer, 2020.

\bibitem{kingma2014adam}
Diederik~P Kingma and Jimmy Ba.
\newblock Adam: A method for stochastic optimization.
\newblock {\em arXiv preprint arXiv:1412.6980}, 2014.

\bibitem{laina2016deeper}
Iro Laina, Christian Rupprecht, Vasileios Belagiannis, Federico Tombari, and
  Nassir Navab.
\newblock Deeper depth prediction with fully convolutional residual networks.
\newblock In {\em 2016 Fourth international conference on 3D vision (3DV)},
  pages 239--248. IEEE, 2016.

\bibitem{lee2019big}
Jin~Han Lee, Myung-Kyu Han, Dong~Wook Ko, and Il~Hong Suh.
\newblock From big to small: Multi-scale local planar guidance for monocular
  depth estimation.
\newblock {\em arXiv preprint arXiv:1907.10326}, 2019.

\bibitem{loshchilov2017decoupled}
Ilya Loshchilov and Frank Hutter.
\newblock Decoupled weight decay regularization.
\newblock {\em arXiv preprint arXiv:1711.05101}, 2017.

\bibitem{MIFDB16}
N. Mayer, E. Ilg, P. H{\"a}usser, P. Fischer, D. Cremers, A. Dosovitskiy, and
  T. Brox.
\newblock A large dataset to train convolutional networks for disparity,
  optical flow, and scene flow estimation.
\newblock In {\em IEEE International Conference on Computer Vision and Pattern
  Recognition (CVPR)}, 2016.

\bibitem{mccormac2017scenenet}
John McCormac, Ankur Handa, Stefan Leutenegger, and Andrew~J Davison.
\newblock Scenenet rgb-d: Can 5m synthetic images beat generic imagenet
  pre-training on indoor segmentation?
\newblock In {\em Proceedings of the IEEE International Conference on Computer
  Vision}, pages 2678--2687, 2017.

\bibitem{Silberman:ECCV12}
Pushmeet~Kohli Nathan~Silberman, Derek~Hoiem and Rob Fergus.
\newblock Indoor segmentation and support inference from rgbd images.
\newblock In {\em ECCV}, 2012.

\bibitem{paszke2019pytorch}
Adam Paszke, Sam Gross, Francisco Massa, Adam Lerer, James Bradbury, Gregory
  Chanan, Trevor Killeen, Zeming Lin, Natalia Gimelshein, Luca Antiga, et~al.
\newblock Pytorch: An imperative style, high-performance deep learning library.
\newblock {\em Advances in neural information processing systems},
  32:8026--8037, 2019.

\bibitem{qi2018geonet}
Xiaojuan Qi, Renjie Liao, Zhengzhe Liu, Raquel Urtasun, and Jiaya Jia.
\newblock Geonet: Geometric neural network for joint depth and surface normal
  estimation.
\newblock In {\em Proceedings of the IEEE Conference on Computer Vision and
  Pattern Recognition}, pages 283--291, 2018.

\bibitem{ramamonjisoa2019sharpnet}
Michael Ramamonjisoa and Vincent Lepetit.
\newblock Sharpnet: Fast and accurate recovery of occluding contours in
  monocular depth estimation.
\newblock In {\em Proceedings of the IEEE/CVF International Conference on
  Computer Vision Workshops}, pages 0--0, 2019.

\bibitem{Ranftl2020}
Ren\'{e} Ranftl, Katrin Lasinger, David Hafner, Konrad Schindler, and Vladlen
  Koltun.
\newblock Towards robust monocular depth estimation: Mixing datasets for
  zero-shot cross-dataset transfer.
\newblock {\em IEEE Transactions on Pattern Analysis and Machine Intelligence
  (TPAMI)}, 2020.

\bibitem{ranftl2016dense}
Rene Ranftl, Vibhav Vineet, Qifeng Chen, and Vladlen Koltun.
\newblock Dense monocular depth estimation in complex dynamic scenes.
\newblock In {\em Proceedings of the IEEE conference on computer vision and
  pattern recognition}, pages 4058--4066, 2016.

\bibitem{ronneberger2015u}
Olaf Ronneberger, Philipp Fischer, and Thomas Brox.
\newblock U-net: Convolutional networks for biomedical image segmentation.
\newblock In {\em International Conference on Medical image computing and
  computer-assisted intervention}, pages 234--241. Springer, 2015.

\bibitem{roy2016monocular}
Anirban Roy and Sinisa Todorovic.
\newblock Monocular depth estimation using neural regression forest.
\newblock In {\em Proceedings of the IEEE conference on computer vision and
  pattern recognition}, pages 5506--5514, 2016.

\bibitem{santurkar2018does}
Shibani Santurkar, Dimitris Tsipras, Andrew Ilyas, and Aleksander M{\k{a}}dry.
\newblock How does batch normalization help optimization?
\newblock In {\em Proceedings of the 32nd international conference on neural
  information processing systems}, pages 2488--2498, 2018.

\bibitem{saxena2005learning}
Ashutosh Saxena, Sung~H Chung, Andrew~Y Ng, et~al.
\newblock Learning depth from single monocular images.
\newblock In {\em NIPS}, volume~18, pages 1--8, 2005.

\bibitem{song2021monocular}
Minsoo Song, Seokjae Lim, and Wonjun Kim.
\newblock Monocular depth estimation using laplacian pyramid-based depth
  residuals.
\newblock {\em IEEE Transactions on Circuits and Systems for Video Technology},
  2021.

\bibitem{song2015sun}
Shuran Song, Samuel~P Lichtenberg, and Jianxiong Xiao.
\newblock Sun rgb-d: A rgb-d scene understanding benchmark suite.
\newblock In {\em Proceedings of the IEEE conference on computer vision and
  pattern recognition}, pages 567--576, 2015.

\bibitem{song2017semantic}
Shuran Song, Fisher Yu, Andy Zeng, Angel~X Chang, Manolis Savva, and Thomas
  Funkhouser.
\newblock Semantic scene completion from a single depth image.
\newblock In {\em Proceedings of the IEEE Conference on Computer Vision and
  Pattern Recognition}, pages 1746--1754, 2017.

\bibitem{tan2019efficientnet}
Mingxing Tan and Quoc Le.
\newblock Efficientnet: Rethinking model scaling for convolutional neural
  networks.
\newblock In {\em International Conference on Machine Learning}, pages
  6105--6114. PMLR, 2019.

\bibitem{ummenhofer2017demon}
Benjamin Ummenhofer, Huizhong Zhou, Jonas Uhrig, Nikolaus Mayer, Eddy Ilg,
  Alexey Dosovitskiy, and Thomas Brox.
\newblock Demon: Depth and motion network for learning monocular stereo.
\newblock In {\em Proceedings of the IEEE Conference on Computer Vision and
  Pattern Recognition}, pages 5038--5047, 2017.

\bibitem{vaswani2017attention}
Ashish Vaswani, Noam Shazeer, Niki Parmar, Jakob Uszkoreit, Llion Jones,
  Aidan~N Gomez, {\L}ukasz Kaiser, and Illia Polosukhin.
\newblock Attention is all you need.
\newblock In {\em Advances in neural information processing systems}, pages
  5998--6008, 2017.

\bibitem{wang2019irs}
Qiang Wang, Shizhen Zheng, Qingsong Yan, Fei Deng, Kaiyong Zhao, and Xiaowen
  Chu.
\newblock Irs: A large naturalistic indoor robotics stereo dataset to train
  deep models for disparity and surface normal estimation.
\newblock {\em arXiv preprint arXiv:1912.09678}, 2019.

\bibitem{wang2019anytime}
Yan Wang, Zihang Lai, Gao Huang, Brian~H Wang, Laurens Van Der~Maaten, Mark
  Campbell, and Kilian~Q Weinberger.
\newblock Anytime stereo image depth estimation on mobile devices.
\newblock In {\em 2019 International Conference on Robotics and Automation
  (ICRA)}, pages 5893--5900. IEEE, 2019.

\bibitem{yin2019enforcing}
Wei Yin, Yifan Liu, Chunhua Shen, and Youliang Yan.
\newblock Enforcing geometric constraints of virtual normal for depth
  prediction.
\newblock In {\em Proceedings of the IEEE/CVF International Conference on
  Computer Vision}, pages 5684--5693, 2019.

\bibitem{yu2015multi}
Fisher Yu and Vladlen Koltun.
\newblock Multi-scale context aggregation by dilated convolutions.
\newblock {\em arXiv preprint arXiv:1511.07122}, 2015.

\bibitem{zheng2018t2net}
Chuanxia Zheng, Tat-Jen Cham, and Jianfei Cai.
\newblock T2net: Synthetic-to-realistic translation for solving single-image
  depth estimation tasks.
\newblock In {\em Proceedings of the European Conference on Computer Vision
  (ECCV)}, pages 767--783, 2018.

\end{thebibliography}

}

\appendix

\section{Supplementary}

\subsection{More comparisons on Vari dataset}

\begin{figure*}[!ht]
    \centering
    \includegraphics[width=1\linewidth]{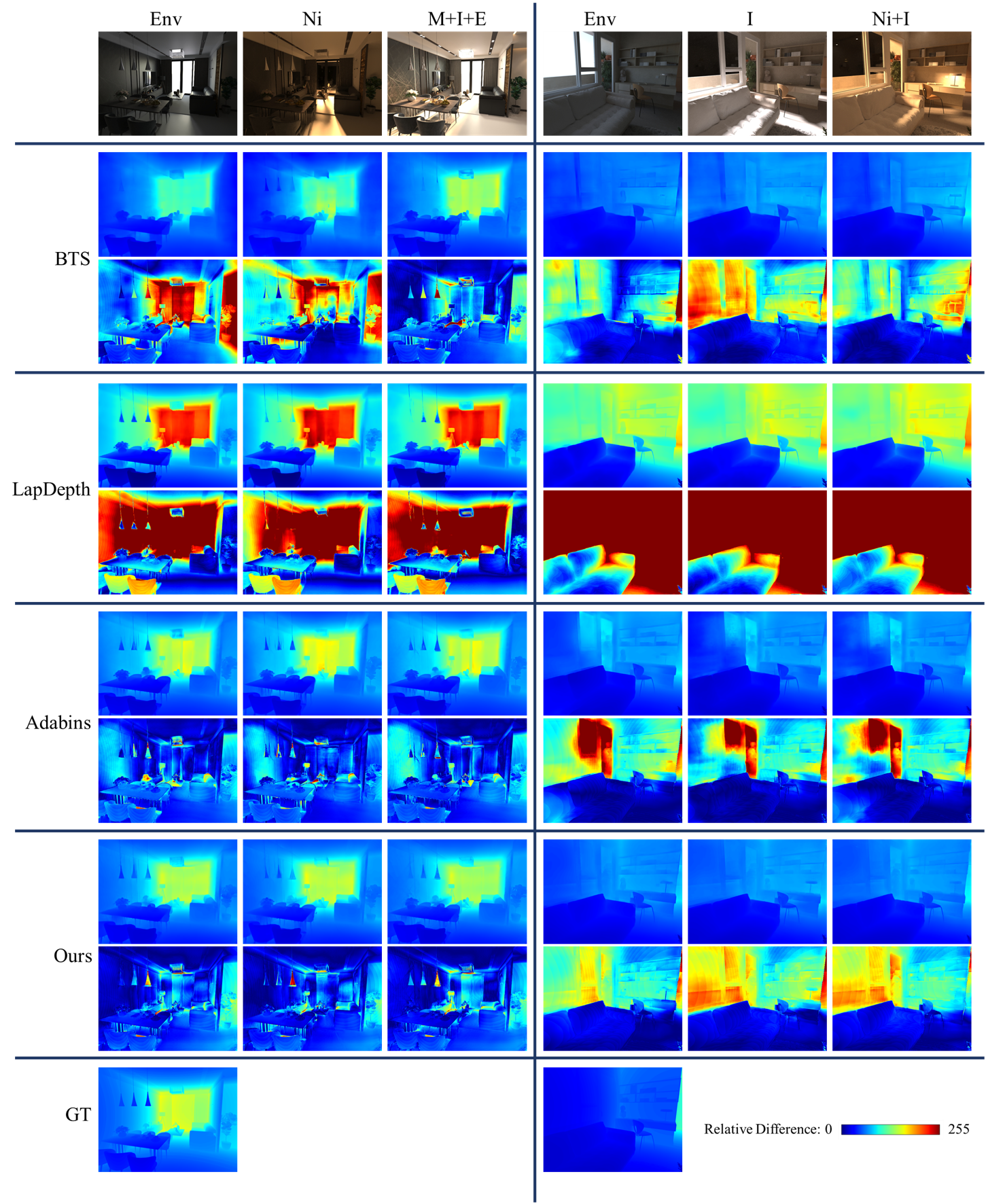}
    \caption{Comparisons on Vari dataset. The first row of each method is the predicted depth map, and the second is the relative difference between this map and the ground truth.}
    \label{fig:more results on vari} 
\end{figure*}

As shown in ~\cref{fig:more results on vari}, we illustrate some comparisons with existing methods under three illuminations on our Vari dataset. We also compute the relative difference between the predicted map and the ground truth and visualize it. We find that if the illumination changes drastically, the depth predicted by other methods differs greatly from the ground truth, especially in the translucent areas such as glasses. On the contrary, our method is able to predict depth maps consistently.

\subsection{More results on real data}

\begin{figure*}[!ht]
    \centering
    \includegraphics[width=1\linewidth]{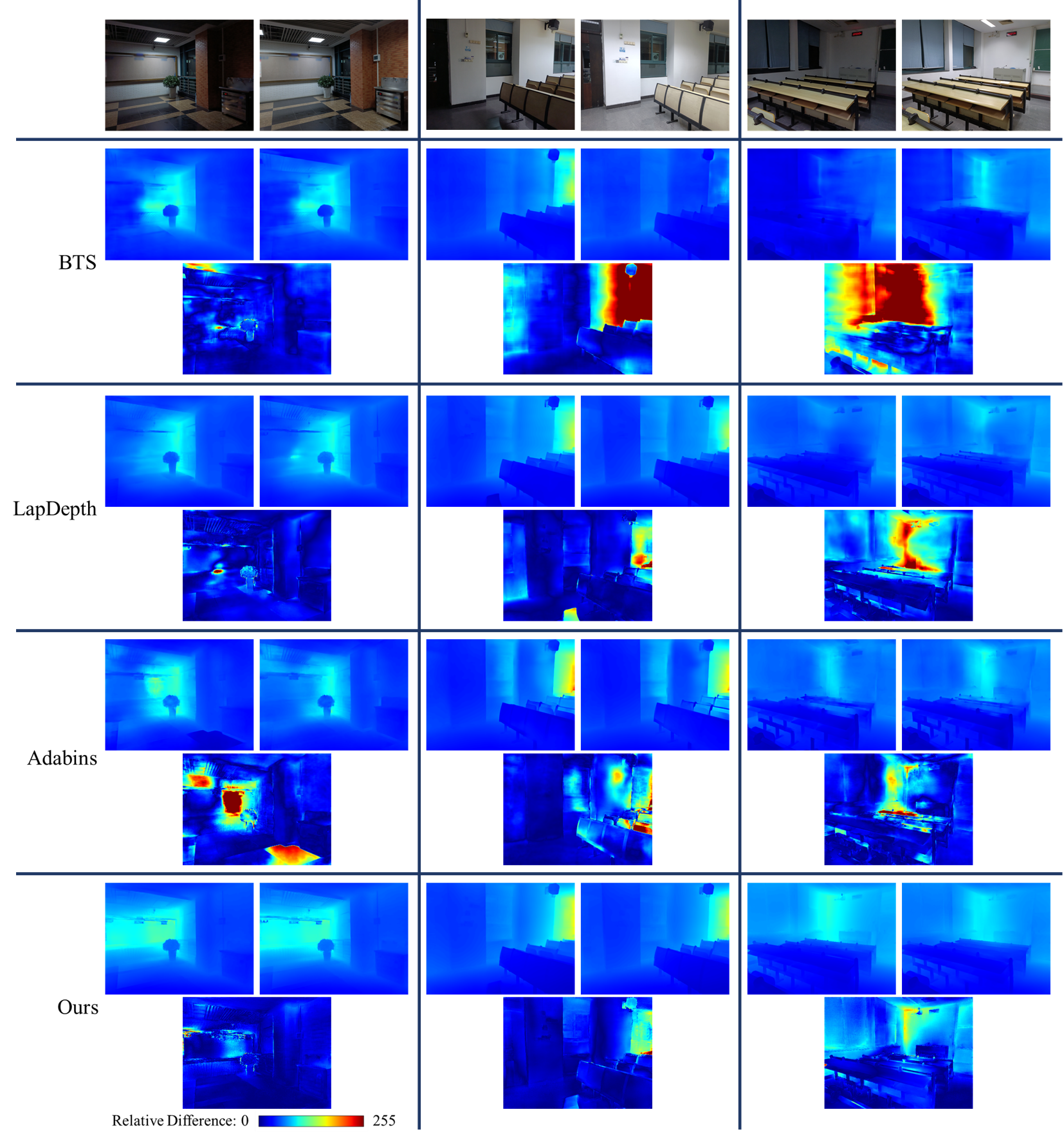}
    \caption{More results on real data. The first row of each method is the predicted depth map and the second is the relative difference under these two illuminations.}
    \label{fig:more results on real 1}
\end{figure*}

\begin{figure*}[!ht]
    \centering
    \includegraphics[width=1\linewidth]{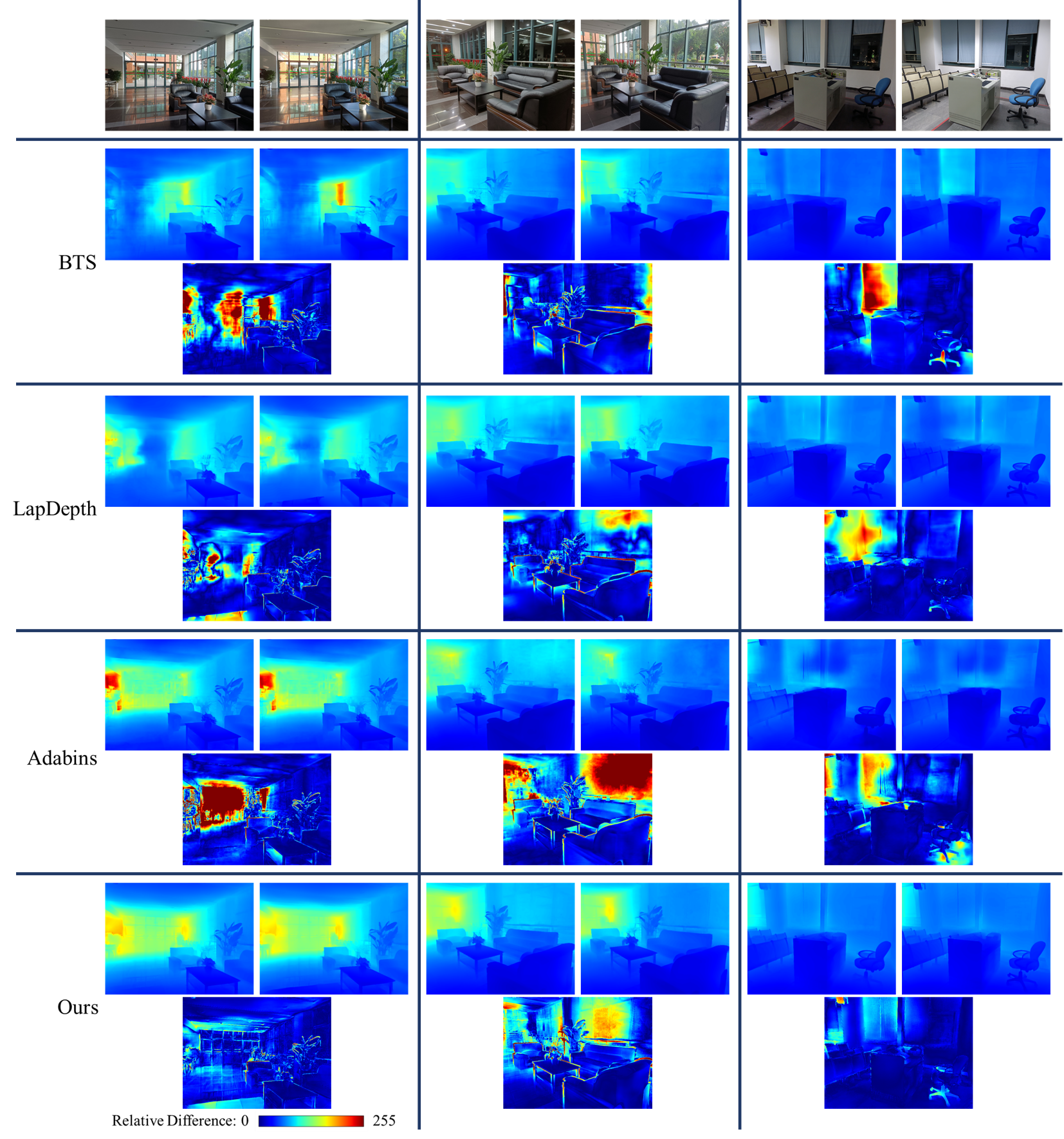}
    \caption{More results on real data. The first row of each method is the predicted depth map and the second is the relative difference under these two illuminations.}
    \label{fig:more results on real 2}
\end{figure*}

\cref{fig:more results on real 1} and ~\cref{fig:more results on real 2} show the comparisons in six real scenes. Similar to the results on our Vari dataset, existing state-of-the art methods fail to estimate consistent and accurate depth values in real scenes under different illuminations. And it also emphasizes that our DCA module makes an important contribution to the generalization and performance of our model.

\subsection{Qualitative results on NYU Depth V2}

\begin{figure*}[!ht]
    \centering
    \includegraphics[width=1\linewidth]{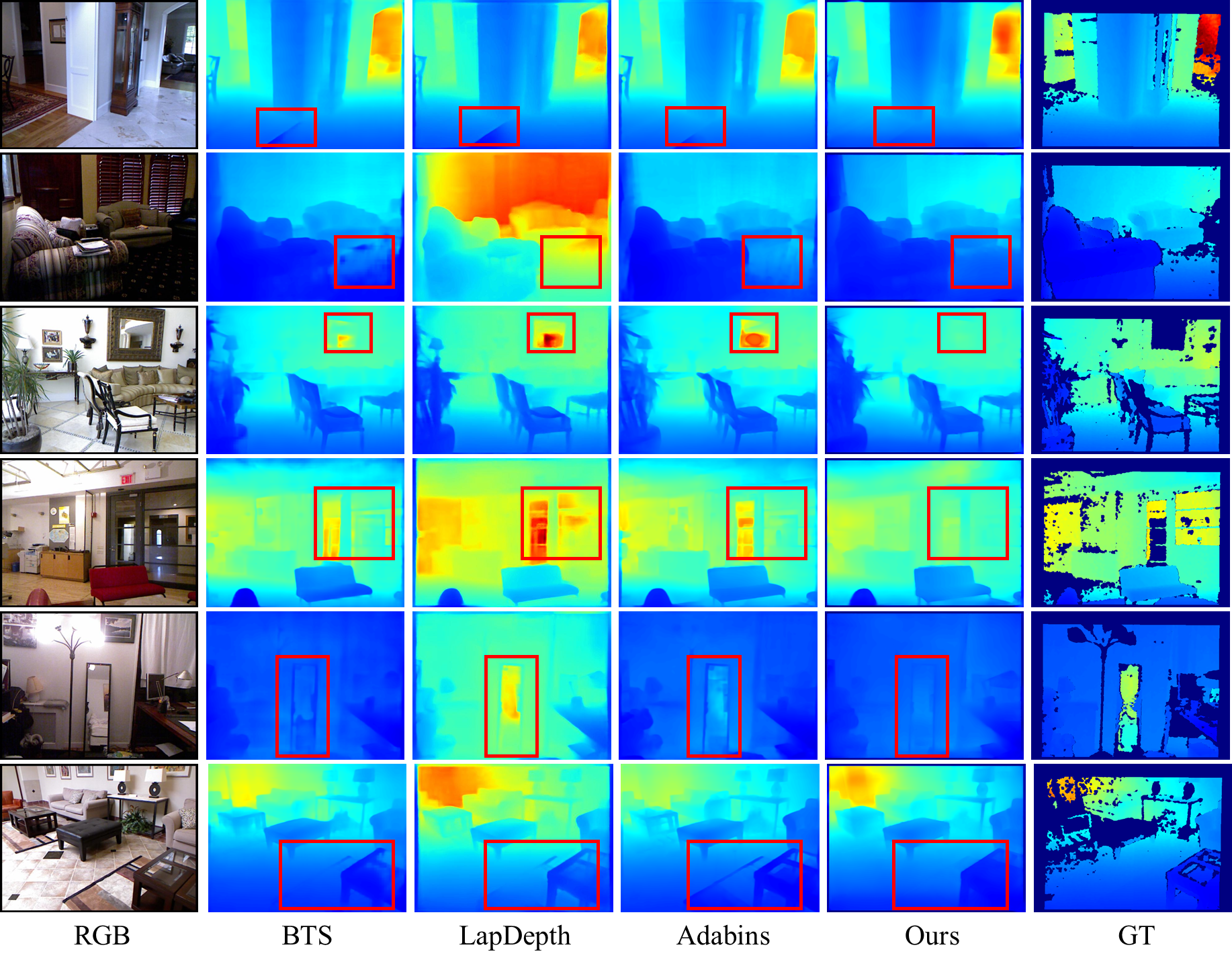}
    \caption{Qualitative results on NYU Depth V2.}
    \label{fig:qualitative results on nyu}
\end{figure*}

\cref{fig:qualitative results on nyu} shows the qualitative results on NYU Depth V2 \cite{Silberman:ECCV12} dataset. As shown in the red rectangles, it can be seen that our DCA module can efficiently exclude depth-irrelevant factors and is able to predict the correct depth values of translucent and reflective areas such as glasses and mirrors even if the ground truth is wrong. Our method also shows a great performance handling different colors on the same plane where other methods mistake them as edges and yield unexpected depth values.

\subsection{Summary}
We compare our method with existing methods on our synthetic Vari dataset, real data, and the popular public dataset NYU Depth V2. We find that existing methods fail to deal with illumination changes and complex layouts, while our method can produce more consistent depth prediction under different illuminations, and is able to predict the accurate depth values for translucent and reflective objects.

\end{document}